\pdfoutput=1

\documentclass[11pt]{article}

\usepackage[]{ACL2023}

\usepackage{times}
\usepackage{latexsym}

\usepackage[T1]{fontenc}

\usepackage[utf8]{inputenc}

\usepackage{microtype}

\usepackage{inconsolata}

\usepackage{booktabs}
\usepackage{diagbox}
\usepackage{xcolor,colortbl}
\usepackage{multirow}
\usepackage{graphicx}
\usepackage{color}
\usepackage{amsmath,bm}
\usepackage{enumitem}

\definecolor{gray}{gray}{0.93}
\definecolor{mygreen}{RGB}{89, 206, 143}
\definecolor{myred}{RGB}{230, 72, 72}
\title{\texttt{FolkScope}: Intention Knowledge Graph Construction for \\  E-commerce  Commonsense Discovery}

\author{Changlong Yu$^{1}$\thanks{~~~Work done during internship at Amazon.} , Weiqi Wang$^{1}$, Xin Liu$^{1}$\footnotemark[1] , Jiaxin Bai$^{1}$\footnotemark[1] , Yangqiu Song$^{1}$\thanks{~~~Visiting academic scholar at Amazon.}\\
\textbf{Zheng Li}$^{2}$, \textbf{Yifan Gao}$^{2}$, \textbf{Tianyu Cao}$^{2}$, \textbf{Bing Yin}$^{2}$ \\
$^{1}$The Hong Kong University of Science and Technology, Hong Kong SAR, China\\ 
$^{2}$Amazon.com Inc, Palo Alto, USA\\ 
\texttt{\{cyuaq, wwangbw, xliucr, jbai, yqsong\}@cse.ust.hk} \\
\texttt{\{amzzhe, yifangao, caoty, alexbyin\}@amazon.com} \\
}

\begin{document}
\maketitle
\begin{abstract}
Understanding users' intentions in e-commerce platforms requires commonsense knowledge.
In this paper, we present \texttt{FolkScope}, an intention knowledge graph construction framework to reveal the structure of humans' minds about purchasing items. 
As commonsense knowledge is usually ineffable and not expressed explicitly, it is challenging to perform information extraction.
Thus, we propose a new approach that leverages the generation power of large language models~(LLMs) and human-in-the-loop annotation to semi-automatically construct the knowledge graph.
LLMs first generate intention assertions via e-commerce-specific prompts to explain shopping behaviors, where the intention can be an open reason or a predicate falling into one of 18 categories aligning with ConceptNet, e.g., {\it IsA}, {\it MadeOf}, {\it UsedFor}, etc.
Then we annotate plausibility and typicality labels of sampled intentions as training data in order to populate human judgments to all automatic generations.
Last, to structurize the assertions, we propose pattern mining and conceptualization to form more condensed and abstract knowledge.
Extensive evaluations and studies demonstrate that our constructed knowledge graph can well model e-commerce knowledge and have many potential applications.
Our codes and datasets are publicly available at \href{https://github.com/HKUST-KnowComp/FolkScope}{https://github.com/HKUST-KnowComp/FolkScope}.
\end{abstract}

\section{Introduction}

In e-commerce platforms, understanding users' searching or purchasing intentions can benefit and motivate a lot of recommendation tasks~\cite{dai2006detecting,zhang2016mining,hao2022dy}.
Intentions are mental states where agents or humans commit themselves to actions.
Understanding others' behaviors and mental states requires rationalizing intentional actions~\cite{sep-folkpsych-theory}, where we need commonsense, or, in other words, good judgements~\cite{liu2004conceptnet}. 
For example, ``at a birthday party, we usually need a birthday cake.''
Meanwhile, commonsense knowledge can be {\it factoid}~\cite{DBLP:conf/aaai/GordonDS10}, which is not invariably true, and is usually ineffable and not expressed explicitly.
Existing intention-based studies on recommendation are either of limited numbers of intention categories~\cite{dai2006detecting,zhang2016mining} or using models to implicitly model the intention memberships~\cite{hao2022dy}.
Thus, it is very challenging to acquire fine-grained intention knowledge in a scalable way.

Existing related knowledge graphs (KGs) can be categorized into two folds.
First, some general situational commonsense KGs deal with everyday social situations~\cite{rashkin-etal-2018-event2mind,sap2019atomic,zhang2020aser}, but they are not directly related to massive products on e-commerce platforms and thus not generalized well on users' behavior data even for generative models, e.g., COMET~\cite{bosselut-etal-2019-comet}.
Second, most e-commerce KGs leverage existing KGs, such as ConceptNet~\cite{liu2004conceptnet,speer2017conceptnet} and Freebase~\cite{DBLP:conf/sigmod/BollackerEPST08}, to integrate them into the e-commerce catalog data~\cite{li2020alimekg,luo2020alicoco,zalmout2021all,luo2021alicoco2,2022_OpenBG}. 
However, such integration is still based on factual knowledge, such as {\it IsA} and {\it DirectorOf} relations, and does not truly model the commonsense knowledge for purchase intentions.
Although some of these KGs may include information related to space, crowd, time, function, and event, they still fall short of modeling true commonsense knowledge~\cite{luo2021alicoco2}.

\begin{figure}
  \centering
  \includegraphics[width=0.98\linewidth]{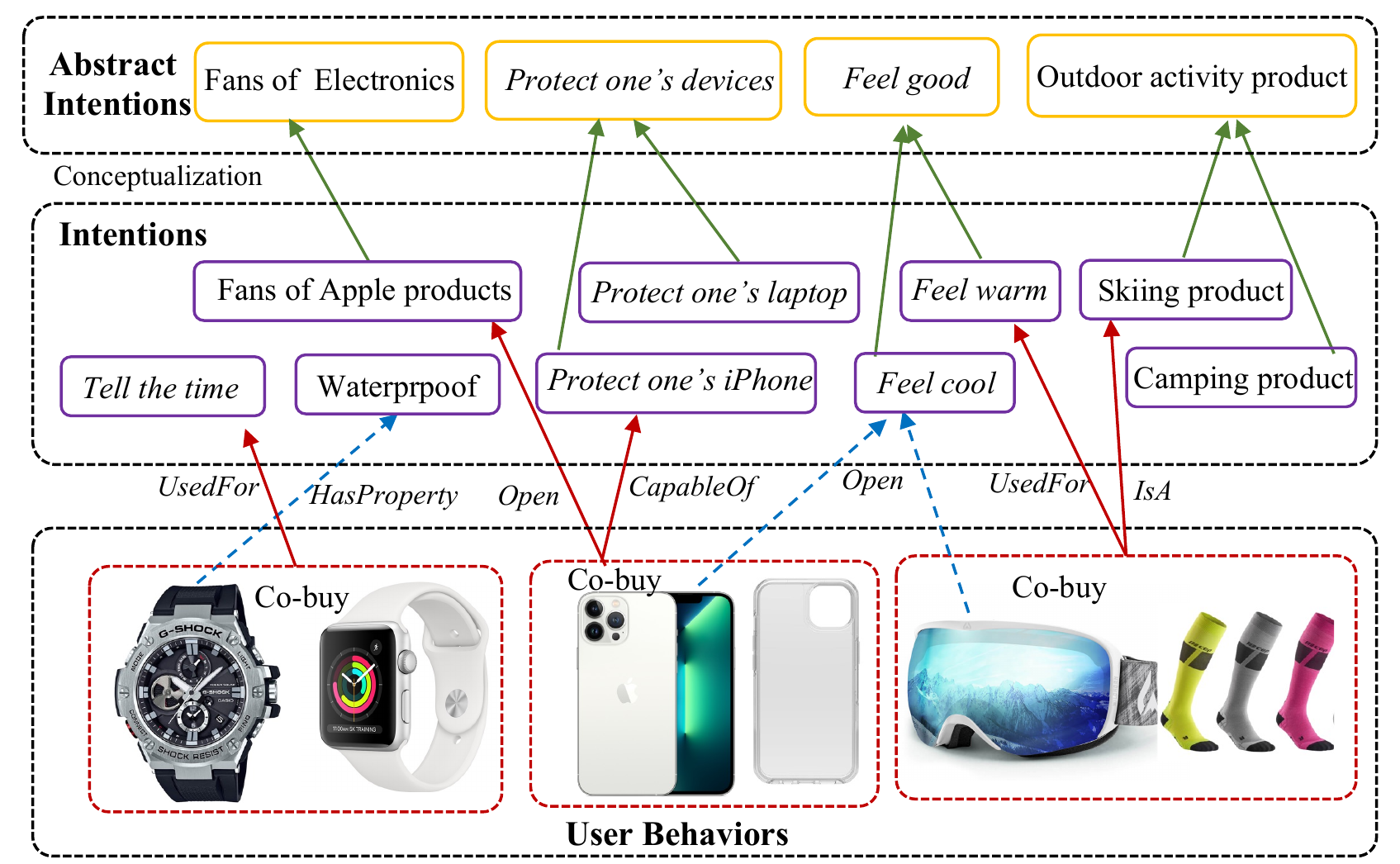} 
  \vspace{-0.1in}
  \caption{An overview of \texttt{FolkScope}. It starts from users' purchasing or co-purchasing behaviors and links them to intentions. Then more abstract intentions are formed to condense the representation of intentions. The intentions can be noun phrases or verb phrases~(\textit{italics}).}
 \label{fig:folkscope}
\end{figure}

Existing KGs constructed for e-commerce platforms can be evaluated for their factual knowledge in terms of \textit{plausibility}.
However, when it comes to purchasing intentions, a person's beliefs and desires~\cite{kashima1998category} are mediated by their intentions, which can be reflected by the {\it typicality} of commonsense~\cite{chalier2020joint,wilhelm2022typical}.
For example, in Figure~\ref{fig:folkscope}, a user bought an Apple watch because ``Apple watches can be used for telling the time'' where the reason is highly plausible (but other watches can also serve similar functions), whereas a more typical reason would be ``apple watches are able to track running,'' or ``the user is simply a fan of Apple products.''
Thus, no matter what kind of factual knowledge a KG contains, if it is not directly linked to rationalization, it cannot be regarded as typical commonsense.
In addition, the task of explaining a user's rating of an item has been proposed as a means of providing recommendations.
To achieve this, researchers have suggested using online reviews as a natural source of explanation~\cite{ni-etal-2019-justifying,DBLP:conf/cikm/LiZC20}.
However, online reviews are often noisy and diverse and may not directly reflect the user's intention behind their purchase or rating.
Instead, they may reflect the consequences of the purchase or the reasons behind the user's rating.
Existing sources of information, such as question-answering pairs, reviews, or product descriptions, do not explicitly mention the user's intentions behind their purchases, making it a challenge to extract intentional commonsense knowledge for e-commerce.
As a result, constructing an intention KG for e-commerce requires sophisticated information extraction techniques and thus remains challenging.

In this paper, we propose a new framework, \texttt{FolkScope}, to acquire intention knowledge in e-commerce. 
Instead of performing information extraction, we start from enormous user behaviors that entail sustainable intentions, such as \textit{co-buy} behaviors, and leverage the generation power of large language models~(LLMs), e.g., GPT~\citep{radford2019language,brown2020language,ouyang2022training}, to generate possible intentions of the purchasing behaviors as candidates.
LLMs have shown the capability of memorizing factual and commonsense knowledge~\cite{petroni2019language,west-etal-2022-symbolic}, and ``sometimes infer approximate, partial representations of the beliefs, desires, and intentions possessed by the agent that produced the context''~\cite{andreas-2022-language}.
As open prompts in the above example can be arbitrary and loosely constrained, we also align our prompts with 18 ConceptNet relations, such as {\it IsA}, {\it HasPropertyOf}, {\it CapableOf}, {\it UsedFor}, etc.
In addition, as the generated knowledge by LLMs can be noisy and may not be able to reflect human's rationalization of a purchasing action, we also perform human annotation for {\it plausibility} and {\it typicality}.

Given generated candidates and annotations to construct the KG, we first perform pattern mining to remove irregular generations.
Then we train classifiers to populate the prediction scores to all generated data.
Finally, for each of the generated intentions, we perform conceptualization to map the key entities or concepts in the intention to more high-level concepts so that we can build a denser and more abstract KG for future generalization.
An illustration of our KG is shown in Figure~\ref{fig:folkscope}.
To assess the overall quality of our KG, we randomly sample populated assertions and estimate their quality.
Furthermore, we demonstrate the quality and usefulness of our KG by using it in a downstream task, CF-based~(collaborative filtering) recommendation.
The contributions of our work can be summarized as follows.

\noindent$\bullet$ We propose a new framework, \texttt{FolkScope}, to construct large-scale intention KG for discovering e-commerce commonsense knowledge.

\noindent$\bullet$ We leverage LLMs to generate candidates and perform two-step efficient annotation on Amazon data with two popular domains, and the process can be well generalized to other domains. 

\noindent$\bullet$ We define the schema of the intention KG aligning with famous commonsense KG, ConceptNet, and populate a large KG based on our generation and annotation with 184,146 items, 217,108 intentions, 857,972 abstract intentions, and 12,755,525 edges (assertions).

\noindent$\bullet$ We perform a comprehensive study to verify the validity and usefulness of our KG. 

\section{Methodology}

\begin{figure}[t]
    \centering
    \includegraphics[width=\linewidth]{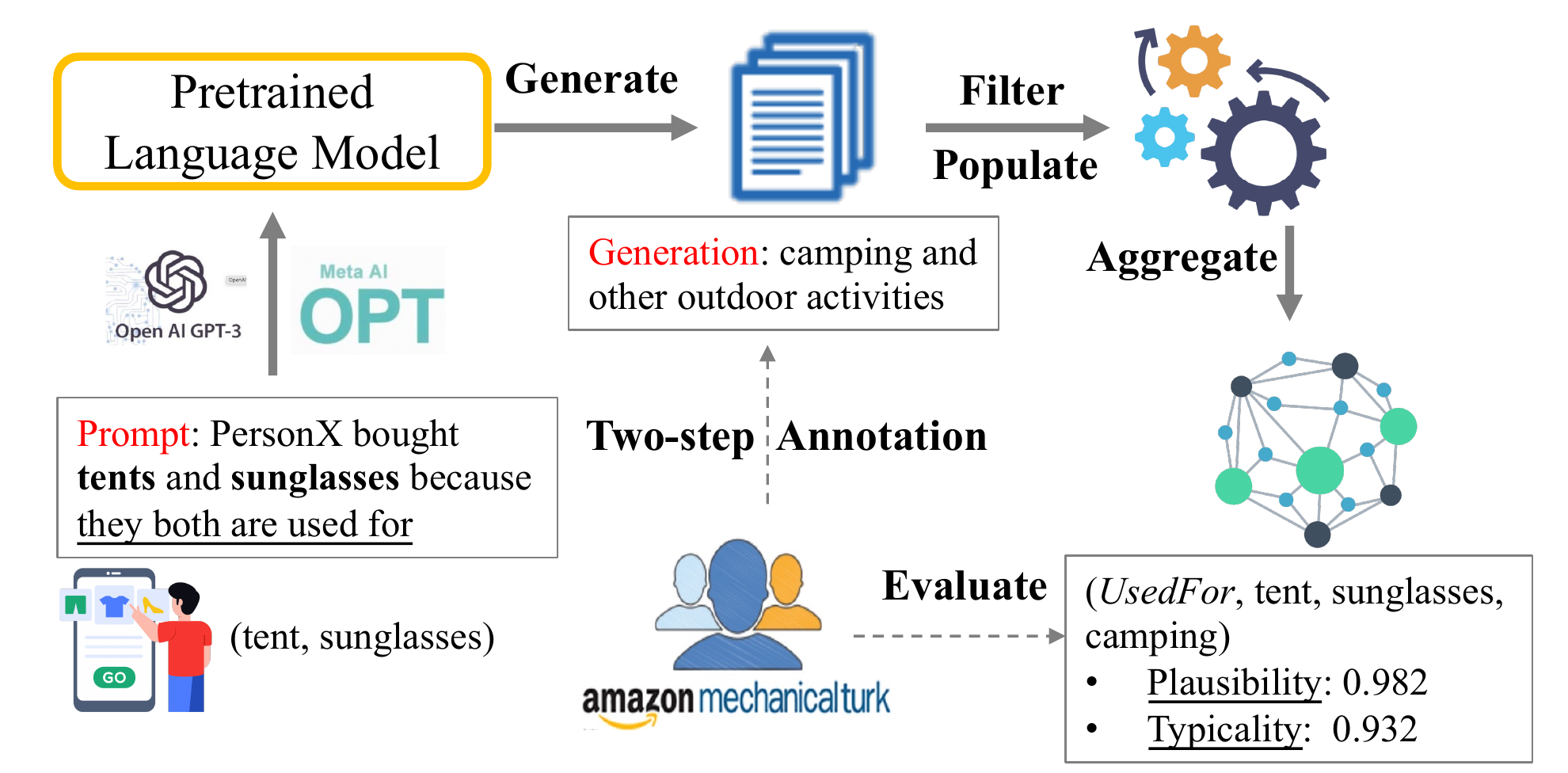}
  \vspace{-0.2in}
    \caption{The overall framework of \texttt{FolkScope}. It includes the generation, population, and conceptualization to semi-automatically construct the e-commerce intention commonsense KG with the help of human-in-the-loop annotations and evaluation.}
    \label{fig:framework}
\end{figure}

\subsection{Overview of \texttt{FolkScope} Framework}

We call our framework \texttt{FolkScope} as we are the first attempt to reveal the structure of e-commerce intentional commonsense to rationalize purchasing behaviors.
As shown in Figure~\ref{fig:framework}, \texttt{FolkScope} is a human-in-the-loop approach for the semi-automatic construction of the KG.
We first leverage the LLMs to generate candidate assertions of intentions for purchasing or co-purchasing behaviors based on \textit{co-buy} data from the released Amazon dataset.
Then we employ two-step annotations to annotate the plausibility and typicality of the generated intentions, where the corresponding definitions of scores are as follows.

\noindent$\bullet$ {\it Plausibility}: how possible the assertion is valid regarding their properties, usages, functions, etc.
    
\noindent$\bullet$ {\it Typicality}: how well the assertion reflects a specific feature that causes the user behavior. Typical intentional assertions should satisfy the following criteria. 
1) {Informativeness:} contains key information about the shopping context rather than a general one, e.g., ``they are used for Halloween parties .'' v.s. ``they are used for the same purpose.'' 
2) {Causality:} captures the typical intention of user behaviors, e.g., ``they have a property of water resistance.'' Some specific attributes or features might largely affect the users' purchase decisions. 

After the annotation, we design classifiers to populate prediction scores to all generated candidates. 
Then the high-quality ones will be further structured using pattern mining on their dependency parses to aggregate similar assertions. 
Then, we also perform conceptualization~\cite{song2011concept,DBLP:journals/ai/ZhangLPKOFS22} to further aggregate assertions to form more abstract intentions.

\subsection{Knowledge Generation}
\label{sec:know_gen}

\noindent \textbf{User Behavior Data Sampling.}
We extract the users' behavior datasets from open-sourced Amazon Review Data~(2018)\footnote{\url{https://nijianmo.github.io/amazon/}}~\citep{ni-etal-2019-justifying} with 15.5M items from Amazon.com. 
In our work, we mainly consider \textit{co-buy} pairs, which might indicate stronger shopping intent signals than \textit{co-view} pairs. 
After the pre-processing and removing duplicated items, the resulting co-buy graph covers 3.5M nodes and 31.4M edges. 
The items are organized into 25 top-level categories from the Amazon website, and among them, we choose two frequent categories: ``\textit{Clothing, Shoes \& Jewelry}" and ``\textit{Electronics}" to sample \textit{co-buy} pairs because those items substantially appear in situations requiring commonsense knowledge to understand, while other categories such as ``Movie'' or ``Music'' are more relevant to factual knowledge between entities. 
We uniformly sample \textit{co-buy} pairs from the two categories, and the statistics are shown in Table~\ref{tab:cobuy_stat}.

\begin{table}[t]
\tiny
\centering
\begin{tabular}{l|ccc}
\toprule
  &  Clothing  &  Electronics & Total\\
\midrule
\# Item Pairs & 199,560 & 93,889 & 293,449 \\
\# Unique Items & 151,509 & 64,244 & 211,349 \\
\# Assertions & 11,358,637 & 5,282,273 & 16,640,910 \\
\# Uniq. Assertions & 2,865,118 & 1,280,259 & 4,063,764 \\
Avg. \# Tokens & 6.66 & 5.25 & 6.21 \\
\bottomrule
\end{tabular}
\caption{Statistics of sampled co-buy pairs and generated candidate assertions. Note that the prompts in the generation are not included in the calculations of assertion lengths.}\label{tab:cobuy_stat}
\end{table}

\noindent \textbf{Prompted Generation.} As shown in Table~\ref{tab:prompts}, we verbalize the prompt templates using the titles of co-buy pairs.
Besides the general prompt (i.e., ``open''), we also align our prompts with 18 relations in ConceptNet highly related to commonsense.
For example, for the relation {\it HasA}, we can design a prompt ``A user bought `item 1' and `item 2' because they both have [GEN]'' where [GEN] is a special token indicating generation.
Since the long item titles might contain noise besides useful attributes, we use heuristic rules to filter out items whose titles potentially affect the conditional generation, like repeated words. 
We use the OPT model~\citep{zhang2022opt} of 30B parameters\footnote{\url{https://huggingface.co/facebook/opt-30b}} with two NVIDIA A100 GPUs based on the HuggingFace library~\citep{wolf-etal-2020-transformers} to generate assertion candidates\footnote{As we will further annotate the plausibility and typicality of candidates, larger models will reduce annotation cost. However, the generation is also constrained by API or computational cost. Thus, we choose the best model we can use.}.
For each relation of the co-buy pairs, we set the max generation length as 100 and generate 3 assertions using nucleus sampling~(p = 0.9)~\citep{holtzman2019curious}. 
We post-process the candidates as follows. (1)~We discard the generations without one complete sentence. (2)~We use the sentence segmenter from Spacy library\footnote{\url{https://spacy.io/}} to extract the first sentence for longer generations. 
After removing duplicates, we obtain 16.64M candidate assertions for 293K item pairs and 4.06M unique tails among them.
The statistics of the two categories are listed in Table~\ref{tab:cobuy_stat}.

\begin{table}[t]\tiny
\centering
\setlength\tabcolsep{4pt}
\begin{tabular}{l|c|c}
\toprule
Type  & Relation &  Prompt       \\ \midrule 
Open &  / & /  \\ \midrule
\multirow{10}{*}{\begin{minipage}{0.4in}Item\end{minipage}}
& \textit{HasA}  & they both have \\
& \textit{HasProperty}  &  they both have a property of     \\
& \textit{RelatedTo}  & they both are related to      \\
& \textit{SimilarTo}  & they both are similar to      \\
& \textit{PartOf}  & they both are a part of    \\
& \textit{IsA}  & they both are a type of      \\
& \textit{MadeOf}  & they both are made of      \\
& \textit{CreatedBy}  & they are created by  \\
& \textit{DistinctFrom}  & they are distinct from  \\
& \textit{DerivedFrom}  & they are derived from  \\
\midrule
\multirow{5}{*}{\begin{minipage}{0.4in}Function\end{minipage}}
& \textit{UsedFor}  &  they are both used for    \\
& \textit{CapableOf}  &   they both are capable of    \\
& \textit{SymbolOf}  &   they both are symbols of  \\
& \textit{MannerOf}  &    they both are a manner of   \\
& \textit{DefinedAs}  &  they both are defined as \\
\midrule
\multirow{3}{*}{\begin{minipage}{0.4in}Human\end{minipage}}
& \textit{Result}  & as a result, the person   \\
& \textit{Cause}  &  the person wants to     \\
& \textit{CauseDesire}  & the person wants his     \\
\bottomrule
\end{tabular}
\caption{Prompts for different commonsense relations. 
}\label{tab:prompts}
\end{table}

\subsection{Two-step Annotation and Population} 
\label{sec:two_round_annotations_and_populations}

As the generated candidates can be noisy or not rational, we apply the human annotation to obtain high-quality assertions and then populate the generated assertions. We use Amazon Mechanical Turk~(MTurk) to annotate our data. Annotators are provided with a pair of co-buy items with each item's title, category, shopping URL, and three images from our sampled metadata. Assertions with different relations are presented in the natural language form by using the prompts presented in Table~\ref{tab:prompts}. More details are listed in Appendix~\ref{sec:annotation_detail}. 
%

\noindent \textbf{Annotation.}
To filter out incorrect candidates, we begin by annotating plausibility in the first step.
This step serves as a preliminary filter and reduces the annotation cost for the subsequent steps.
We randomly sample 66K generations and collect three plausibility votes per generated candidate.
The final plausibility score is derived by majority voting. 
The overall IAA score is 75.48\% in terms of pairwise agreement proportion, while Fleiss's Kappa~\cite{fleiss1971measuring} is 0.4872.
Both metrics are satisfiable for such large-scale annotations.

Different from the simple binary plausibility judgments, in the second step, we have more fine-grained and precise typicality indicators concerning \textit{informativeness} and \textit{causality}. Here we choose the candidates automatically labeled as plausible based on our classifier trained on the first step's data. We ask the annotators to judge whether they are \textit{strongly acceptable} (+1), \textit{weakly acceptable} (0.5), {\it rejected} (0), or \textit{implausible} (-1) that the assertion is informative and casual for a purchasing behavior. 
Considering the judgments might be subjective and biased with respect to different annotators, we collect five annotations for each assertion and take the average as the final typicality score.\footnote{The annotators in this step are chosen from the high-quality annotators in the first step. We tried other options, such as using seven or nine annotators per generation in our pilot study. The results do not show much improvement.}
Similar to the first step, we collect around 60K assertions.
Empirically, we find annotating more data does not bring significantly better filtering accuracy.
The statistics are presented in Table~\ref{tab:labled_stat}. 

\begin{table}[t]\tiny
\centering
\setlength\tabcolsep{4pt}
    \begin{tabular}{l|c|cc}
    \toprule
     \multicolumn{1}{c|}{Stage}  & \multicolumn{1}{c|}{Category}  &  \multicolumn{1}{c}{\# Annotation}    &  \multicolumn{1}{c}{Avg. Score}    \\ \midrule 
    \multirow{3}{*}{Plausibility} & Clothing  &  44,337 & 0.6435 \\
    & Electronics  &  21,760 & 0.5467 \\
    & Total  & 66,097 & 0.6116 \\
    \midrule
    \multirow{3}{*}{Typicality} & Clothing & 38,279 & 0.4407 \\
    & Electronics & 22,995 & 0.4631 \\
    & Total & 61,274 & 0.4491 \\
    \bottomrule
    \end{tabular}
    \vspace{-0.05in}
\caption{Statistics of annotated data.}\label{tab:labled_stat}
\end{table}

\begin{table}[t]\tiny
\centering
\setlength\tabcolsep{4pt}
    \begin{tabular}{l|ccc}
    \toprule
       & Plausibility  &  Typicality   \\ \midrule 
    RoBERTa-large & 83.22\% & 81.96\% \\
    DeBERTa-large & 85.12\% & 82.67\% \\
    \bottomrule
    \end{tabular}
    \vspace{-0.05in}
\caption{Classification results on validation sets (F1).}\label{tab:labled_classification}
\end{table}

\noindent \textbf{Population.} 
For plausibility population, we train binary classifiers based on the majority voting results in the first step, which can produce binary labels of the plausibility of unverified generations.
For the typicality score, as we take the average of five annotators as the score, we empirically use scores greater than 0.8 to denote positive examples and less than 0.2 as negative examples.
We split the train/dev sets at the ratio of 80\%/20\% and train binary classifiers using both DeBERTa-large~\citep{he2021deberta,he2023debertav} and RoBERTa-large~\citep{liu2019roberta} as base models.
The best models are selected to maximize the F1 scores on the validation sets, and results are shown in Table~\ref{tab:labled_classification} (more results can be found in Appendix~\ref{appendix:KBP}). 
DeBERTa-large achieves better performance than RoBERTa-large on both plausibility and typicality evaluation. 
We populate the inference over the whole generated corpus in Table~\ref{tab:cobuy_stat} and only keep the assertions whose predicted plausibility scores are above 0.5~(discarding 32.5\% generations and reducing from 16.64M to 11.24M). 
Note that only plausible assertions are kept in the final KG.
Using different confidence cutting-off thresholds leads to trade-offs between the accuracy of generation and the size of the corpus. 
After the two-step populations, we obtain the plausibility score and typicality score for each assertion. Due to the measurement of different aspects of knowledge, we observe low correlations between the two types of scores~(Spearman correlation $\rho$: 0.319 for \textit{clothing} and 0.309 for \textit{electronics}).

\subsection{Knowledge Aggregation}
\label{sec:know_constct}

\begin{figure}[t]
    \centering
    \includegraphics[width=1\linewidth]{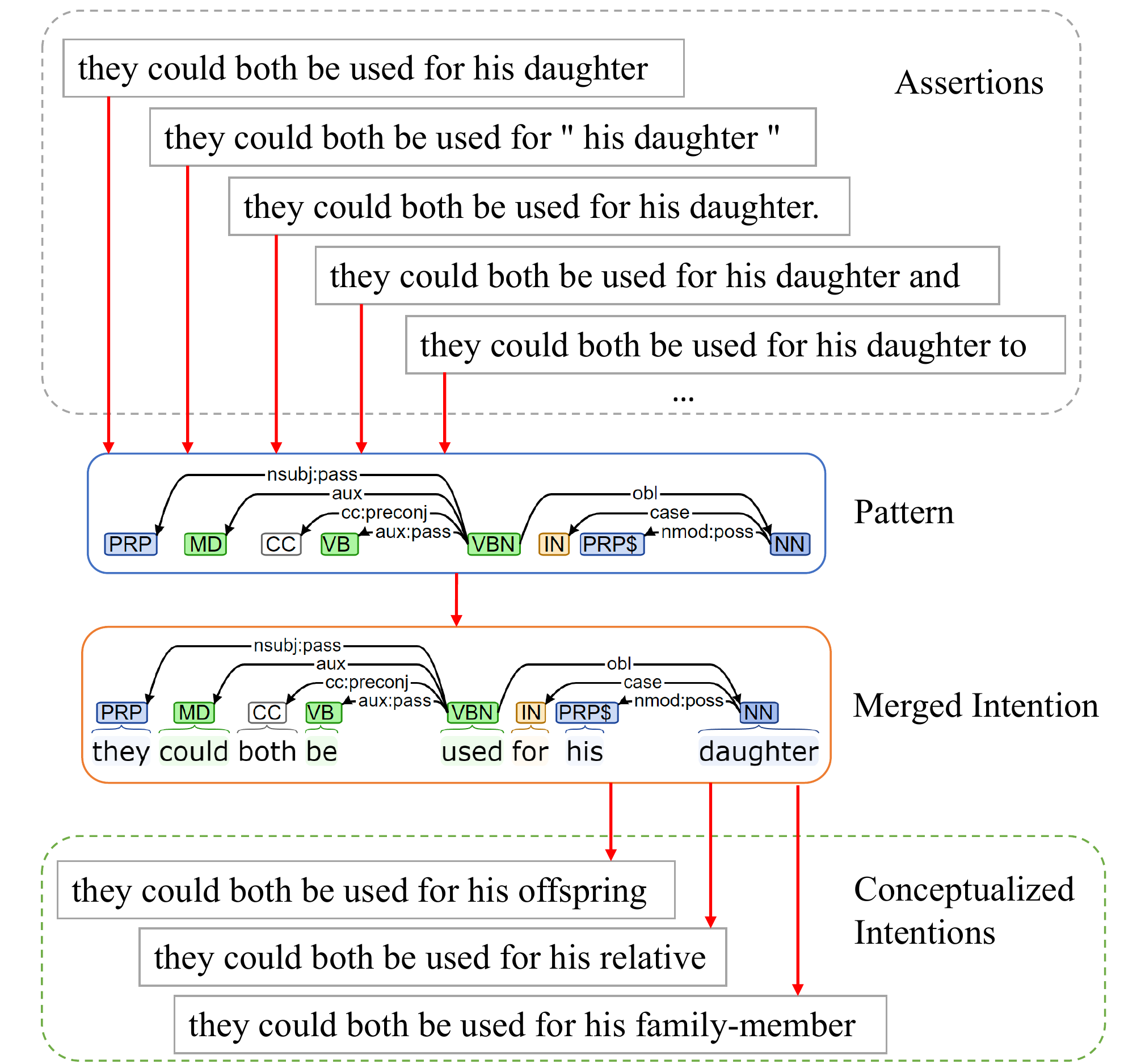}
  \vspace{-0.05in}
    \caption{Illustration of knowledge aggregation.}
    \label{fig:know_agg}
\end{figure}

To acquire a KG with topology structures instead of sparse triplets, we aggregate semantically similar assertions.
This is done by (1) pattern mining to align similar generated patterns and (2) conceptualization to produce more abstract knowledge.

Assertions are typically expressed as free-form text phrases, some of which may have similar syntax and semantics.
By extracting the skeleton and necessary modifiers, such as demonstrative pronouns, adjectives, and adverbs, we can reduce the noise generated by these phrases.
For example, as shown in Figure~\ref{fig:know_agg}, several generations can be simplified to ``they could both be used for his daughter,'' despite the presence of punctuation and incomplete content.
To achieve this, we employ frequent graph substructure mining over dependency parse trees to discover linguistic patterns~(More details in Appendix~\ref{appendix:pattern_mining}).

After pattern mining, we can formally construct our knowledge graph, where the head is a pair of items $(p_1, p_2)$, the relation $r$ is one of the relations shown in Table~\ref{tab:prompts}, and the tail is an aggregated assertion $e$ that is originally generated and then mapped to a particular one among 256 patterns.
Each of the knowledge triples is associated with two populated scores, i.e., plausibility and typicality.

To produce abstract knowledge generalizable to new shopping contexts, we also consider the conceptualization with the large-scale concept KG, Probase~\cite{wu2012probase,he2022acquiring,wang2023cat}.
The conceptualization process maps one extracted assertion $e$ to multiple conceptualized assertions with concepts $c$.
For example, in Figure~\ref{fig:know_agg}, ``they could be used for his daughter'' can be conceptualized as ``they could be used for his offspring,'' ``they could be used for his relative,'' and ``they could be used for his family-member,'' etc. 
The conceptualization weight $P(c|e)$ can be determined by the likelihood for $\rm{IsA}(e, c)$ in Probase.
This process has been employed and evaluated by ASER 2.0~\cite{DBLP:journals/ai/ZhangLPKOFS22}.
Finally, we obtain a KG with 184,146 items, 217,108 intentions, 857,972 abstract intentions, and 12,755,525 edges to explain 236,739 co-buy behaviors, where 2,298,011 edges from the view of original assertions and 9,297,500 edges from the angle of conceptualized ones, and 1,160,014 edges model the probabilities of the conceptualization.

\begin{table}[t]
\tiny
\setlength\tabcolsep{5pt}
\centering
\begin{tabular}{c|cc|cc|cc}
\toprule
\multirow{2}{*}{Threshold}& \multicolumn{2}{c|}{{\bf Clothing}}  & \multicolumn{2}{c|}{{\bf Electronics}}& \multicolumn{2}{c}{{\bf Total}} \\
& Accept & Size & Accept  & Size & Accept & Size\\ \midrule
0.5 & 83.73\%  & 7,986,031 & 82.74\% & 3,250,605 & 83.40\% & 11,236,636 \\
0.7 & 90.27\%  & 7,346,160 & 88.27\% & 2,868,256 & 89.40\% & 10,214,416 \\
0.8 & 91.02\%  & 6,947,606 & 89.50\% & 2,650,625 & 90.00\% & 9,598,231 \\
0.9 & 95.60\%  & 6,167,315 & 94.87\% & 2,230,423 & 95.36\% & 8,397,738 \\
\bottomrule
\end{tabular}
  \vspace{-0.05in}
\caption{Acceptance ratios of plausible assertions and the corresponding sizes of populated assertions with different cutting-off thresholds.}\label{tab:cutoff_stat}
\end{table}

\begin{table}[t]
\tiny
\centering
\begin{tabular}{l|c|ccc}
\toprule
\multicolumn{1}{c|}{Relation} & \multicolumn{1}{c|}{Acc. Rate}  &  \multicolumn{1}{c}{\# Edges} & \multicolumn{1}{c}{\# Tails} & \multicolumn{1}{c}{Avg. Length} \\
\midrule
\textit{Open} & 87.54\% & 703,059 & 151,748 & 7.86 \\
\textit{HasA} & 94.08\% & 710,331 & 68,516 & 5.53 \\
\textit{HasProperty} & 79.13\% & 317,938 & 133,877 & 5.00 \\
\textit{RelatedTo} & 91.89\% & 571,918 & 130,551 & 3.08 \\
\textit{SimilarTo} & 86.35\% & 685,737 & 18,603 & 3.53 \\
\textit{PartOf} & 79.60\% & 674,928 & 114,983 & 4.36 \\
\textit{IsA} & 89.05\% & 591,037 & 98,262 & 3.82 \\
\textit{MadeOf} & 90.05\% & 528,289 & 70,246 & 5.06 \\
\textit{CreatedBy} & 95.15\% & 267,459 & 74,920 & 3.93 \\
\textit{DistinctFrom} & 91.74\% & 861,929 & 80,295 & 4.66 \\
\textit{DerivedFrom} & 85.54\% & 444,131 & 61,696 & 4.90 \\
\textit{UsedFor} & 91.79\% & 630,462 & 45,206 & 2.58 \\
\textit{CapableOf} & 87.73\% & 681,480 & 101,170 & 5.23 \\
\textit{SymbolOf} & 78.04\% & 809,196 & 52,075 & 3.46 \\
\textit{MannerOf} & 89.44\% & 371,892 & 122,829 & 4.38 \\
\textit{DefinedAs} & 85.59\% & 288,411 & 151,986 & 6.31 \\
\textit{Result} & 44.79\% & 568,523 & 166,018 & 8.80 \\
\textit{Cause} & 80.50\% & 696,392 & 185,042 & 7.06 \\
\textit{CauseDesire} & 67.23\% & 833,524 & 155,422 & 5.61 \\
\midrule
\textit{Total} & 83.40\% & 11,236,636 & 1,874,782 & 5.02 \\
\bottomrule
\end{tabular}
  \vspace{-0.05in}
\caption{Evaluation on plausible rate and size of the populated KG. The prompts in the generation are not included in the calculations of assertion lengths.}
\label{tab:evaluation_plausibility}
\end{table}

\section{Intrinsic Evaluations}

In this section, we present some examples of our constructed KG and conduct comprehensive intrinsic evaluations of KG.

\begin{table*}[t]
\tiny
\centering
\begin{tabular}{p{2.5cm} | p{2.5cm} | p{1.5cm} | p{3.5cm} | p{0.6cm} | p{0.6cm} }
\toprule
\multicolumn{1}{c|}{Item 1} & \multicolumn{1}{c|}{Item 2} & \multicolumn{1}{c|}{Relation} & \multicolumn{1}{c|}{Tail} & \multicolumn{1}{c|}{P.} & \multicolumn{1}{c}{T.} \\
\midrule
\multirow{3}{*}{\shortstack{GGS III LCD Screen \\ Protector glass for \\ CANON 5D Mark III  \href{https://www.amazon.com/dp/B008DCK0I4}{(link)}}}
& \multirow{3}{*}{\shortstack{ECC5D3B Secure Grip \\ Camera Case for  \\ Canon 5D Mark III  \href{https://www.amazon.com/dp/B008MAGCZM}{(link)}}}
& \textit{Open} & they can be used for the same purpose & 0.67 & 0.35 \\
& & \textit{HasProperty} & ``easy to install" and ``easy to remove" & 0.80 & 0.85 \\
& & \textit{SimilarTo} & \textcolor{myred}{the product he bought} & 0.95 & 0.09 \\
\multirow{4}{*}{\quad \quad \quad  \parbox[c]{1em}{
    \includegraphics[width=0.4in]{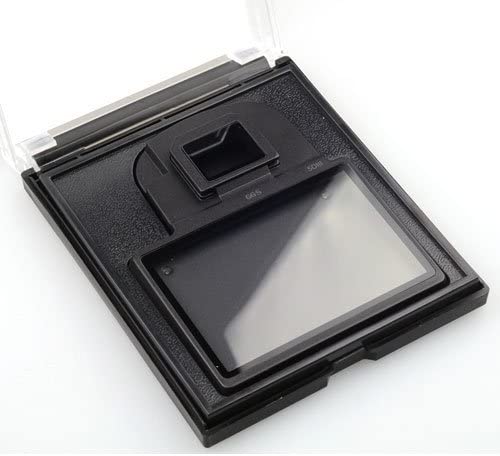}}}
& \multirow{4}{*}{\quad \quad \quad \parbox[c]{1em}{
    \includegraphics[width=0.4in]{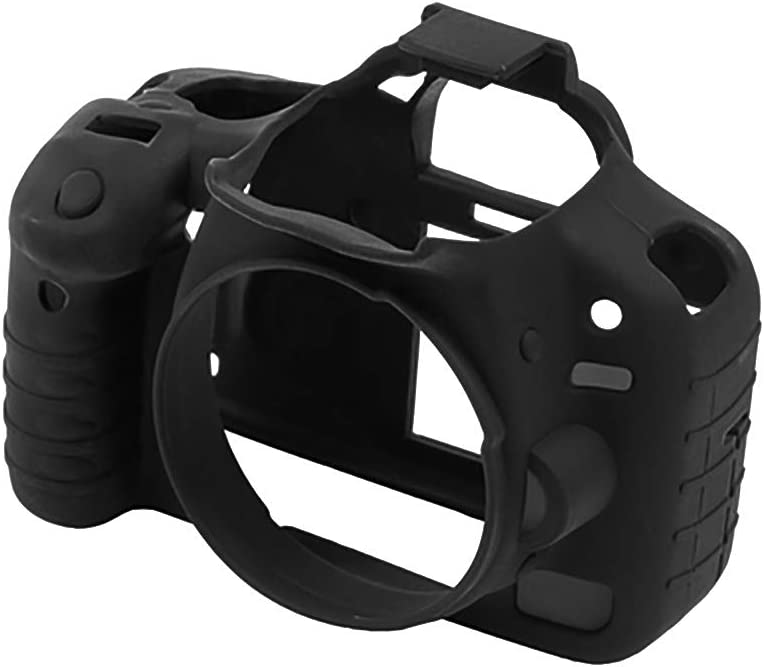}}}
& \textit{PartOf} & \textcolor{mygreen}{his camera gear} & 0.93 & 0.99 \\
& & \textit{UsedFor} & \textcolor{mygreen}{protect the camera from scratches and dust} & 0.97 & 0.99 \\
& & \textit{SymbolOf} &  his love for his camera & 0.99 & 0.88 \\
& & \textit{DefinedAs} & "Camera Accessories" on Amazon.com  & 0.99 & 0.67 \\
\midrule
\multirow{3}{*}{\shortstack{Sun Smarties Baby \\ UPF 50+ Non-Skid Sand \\ Water Socks Pink  \href{https://www.amazon.com/dp/B00ZGSOIM2}{(link)}}}
& \multirow{3}{*}{\shortstack{Schylling UV Play \\ Shade, SPF 50+, \\ Ultra portable, Blue  \href{https://www.amazon.com/dp/B0014I4TYA}{(link)}}}
& \textit{Open} &  \textcolor{mygreen}{he was worried about his baby's skin} & 0.98 & 0.98 \\
& & \textit{SimilarTo} & \textcolor{myred}{each other} & 0.74 & 0.01 \\
& & \textit{DistinctFrom} & \textcolor{myred}{other similar products} & 0.97 & 0.10 \\
\multirow{5}{*}{\quad \quad \quad \parbox[c]{1em}{
    \includegraphics[width=0.4in]{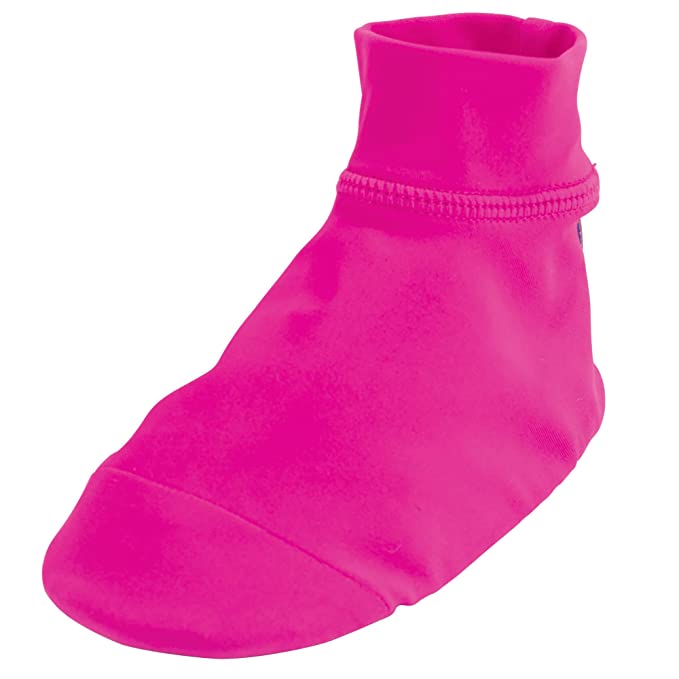}}}
& \multirow{4}{*}{\quad \quad \quad  \parbox[c]{1em}{
    \includegraphics[width=0.4in]{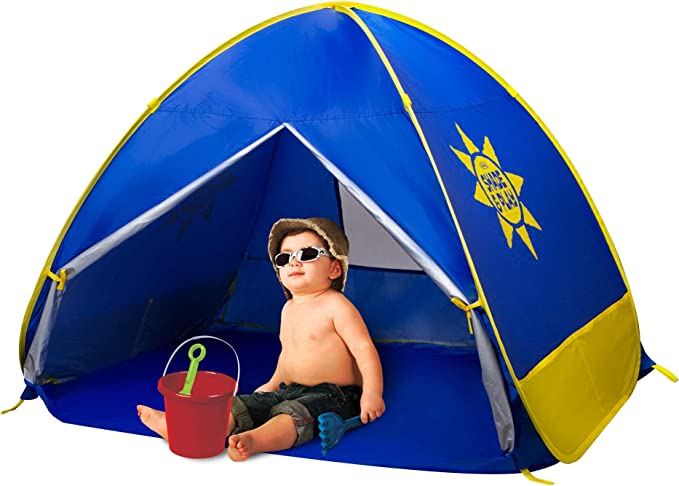}}}
& \textit{UsedFor} & baby's outdoor activities & 0.85 & 0.91 \\
& & \textit{CapableOf} & \textcolor{mygreen}{blocking harmful UV rays} & 0.97 & 0.99 \\
& & \textit{DefinedAs} & sun protection products & 0.87 & 0.81 \\
& & \textit{Result} & \textcolor{mygreen}{enjoy the sun sagely and comfortably} & 0.97 & 0.98 \\
& & \textit{Cause} &  want to use them for his/her baby & 0.99 & 0.94 \\
\bottomrule
\end{tabular}
  \vspace{-0.05in}
\caption{Two examples from the constructed knowledge graph. ``P.'' and ``T.'' stand for the predicted plausibility and typicality scores. Generated tails with high typicality~(in green) and low typicality~(in red) scores are highlighted.}\label{tab:quality_case}

\end{table*}

\subsection{Examples in KG}

We show two examples of co-purchasing products and their corresponding knowledge~($\S$~\ref{sec:know_gen}) as well as populated scores~($\S$~\ref{sec:two_round_annotations_and_populations}) in Table~\ref{tab:quality_case}.
We measure the quality of assertions using both plausibility and typicality scores, which are again shown they are not correlated. 
For example, ``they are {\it SimilarTo} the product they bought'' for the first pair and ``they are {\it DistinctFrom} other similar products'' for the second pair are plausible assertions but not typical explanations of why a user would buy them together.
Moreover, some of the open relations are very good as well.
Take the second pair as an example: the open relation shows ``he was worried about his baby's skin'' as both products are related to baby skin protection.
We also append more typical knowledge examples in Table~\ref{tab:more_typical_examples} of the Appendix.

\subsection{Human Evaluation}\label{sec:human_eval}

As we populate the whole generated assertions using classifiers based on \texttt{DeBERTa-large} model, we conducted human evaluations by sampling a small number of populated assertions from different scales of predicted scores to evaluate the effectiveness of the knowledge population.

\subsubsection{Plausibility Evaluation} We randomly sample 200 plausible assertions from each relation in each of the clothing and electronics domains to test the human \textit{acceptance rate}. 
The annotation is conducted in the same way as the construction step.
As we only annotate assertions predicted to be greater than the 0.5 plausibility score, the IAA is above 85\%, even greater than the one in the construction step.
As shown in Table~\ref{tab:cutoff_stat}, different cutting-off thresholds~(based on the plausibility score by our model) lead to the trade-offs between the accuracy and the KG size. 
Overall, \texttt{FolkScope} can achieve an 83.4\% acceptance rate with a default threshold~(0.5). 
To understand what is filtered, we manually check the generations with low plausibility scores and find that OPT can generate awkward assertions, such as simply repeating the item titles or obviously logical errors regarding corresponding relations. 
Our classifier trained on annotated datasets helps resolve such cases. Using a larger threshold of 0.9, we attain a 95.35\% acceptance rate, a nearly 11.96\% improvement while still keeping above 8M plausible assertions. 
We also report the accuracy in terms of different relations in Table~\ref{tab:evaluation_plausibility}. 
We can observe that assertions concerning the relations of human beings' situations like {\it Cause}, {\it Result}, and {\it CauseDesire} have relatively lower plausibility scores and longer lengths than the relations of items' property, function, etc. This is because there exist some clues about items' knowledge in the item titles, while it is much harder to generate (or guess) implicit human beings' casual reasons using language generation. 

\begin{table}
\centering
\tiny
\begin{tabular}{c|cc}
\toprule
Threshold & Aggregated Knowledge  & Conceptualization \\
\midrule
0.8 & 0.6215 & 0.4571 \\
0.9 & 0.6335 & 0.5567 \\
0.99 & 0.7028 & 0.5775 \\
\bottomrule
\end{tabular}
  \vspace{-0.05in}
\caption{Average annotated typicality scores for assertions after pattern mining and conceptualization with different thresholds of predicted typicality scores.}
\label{tab:quality_conceptualization_vs_eventuality}
\end{table}

\begin{figure}[t]
    \centering
    \includegraphics[width=1\linewidth]{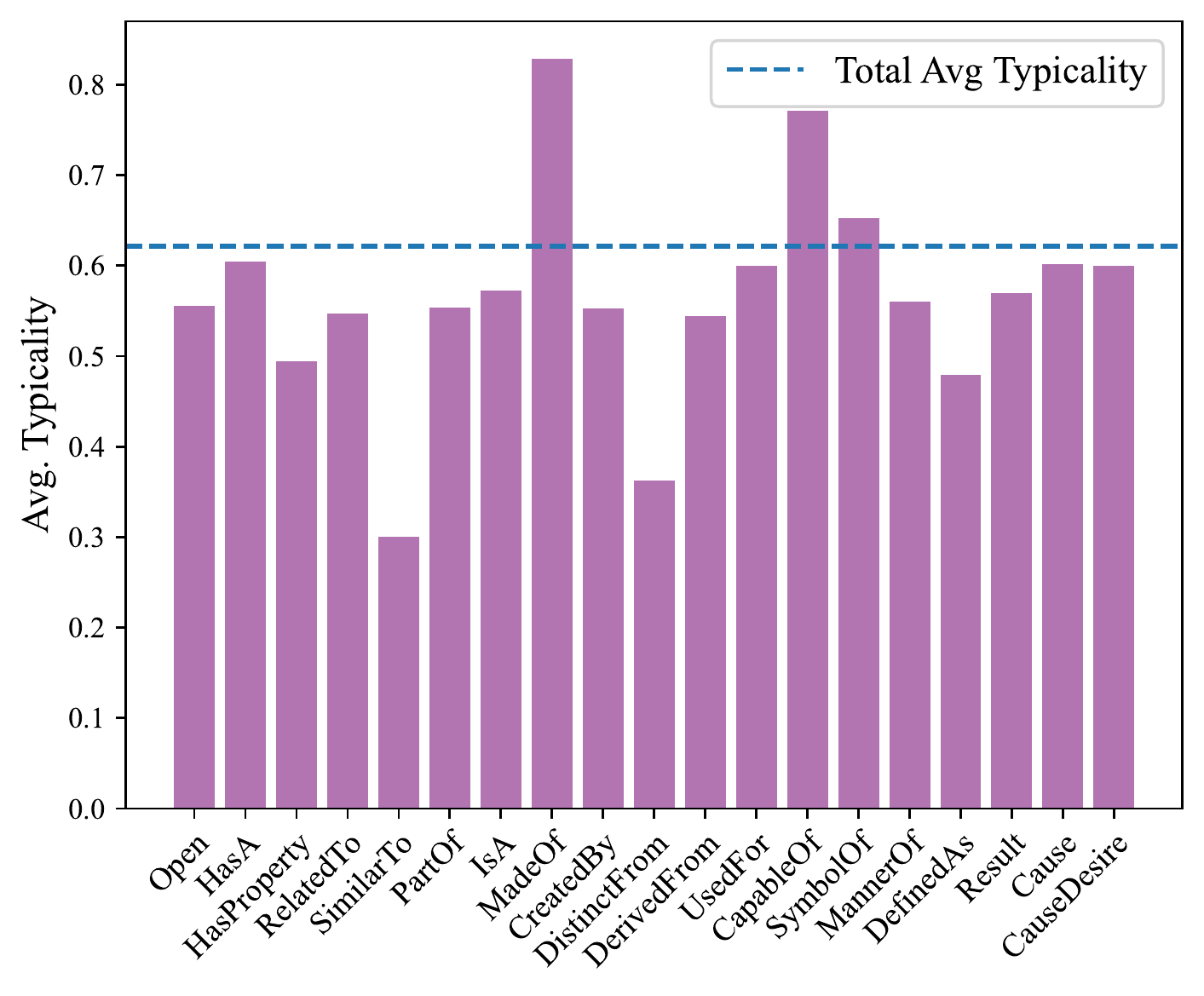}
      \vspace{-0.1in}
    \caption{Average typicality score of each relation in the populated KG with the cutting-off threshold 0.8.}
    \label{fig:quality_evaluation_relation}
\end{figure}

\subsubsection{Typicality Evaluation} The goal of the typicality population is to precisely recognize high-quality knowledge, and we evaluate whether assertions with high typicality scores are truly good ones. We randomly sample 200 assertions from each relation whose predicted typicality scores are above 0.8 for human evaluation. 
Each of the assertions is again annotated by five AMT workers, and the average rating is used.
The results are shown in Table~\ref{tab:quality_conceptualization_vs_eventuality}.
It shows that average annotated scores are lower than the predicted ones due to harder judgments for typicality.
Similarly, predicted typicality scores are less accurate than plausibility.
Especially the typicality score will be further decreased after conceptualization.
This is because, first, the conceptualization model may introduce some noise, and second, the more abstract knowledge tends to be less typical when asking humans to annotate.
We also show the typicality scores of each relation in  Figure~\ref{fig:quality_evaluation_relation}.
Different from plausibility, {\it SimilarTo}, {\it DistinctFrom},  {\it DefinedAs}, and {\it HasPropertyOf} are less typical compared to other relations.
They describe items' general features but can not well capture typical purchasing intentions though they have high plausibility scores,
whereas {\it CapableOf} and {\it MadeOf} are the most typical features that can explain purchasing intentions for the two domains we are concerned about.

More evaluation on the diversity of implicit generation and fine-grained subcategory knowledge aggregation can be found in Appendix~\ref{appendix:more_evaluation}.

\section{Extrinsic Evaluation}

\subsection{Experimental Setup}

\paragraph{Data Preparation.}
We conduct extrinsic evaluation via knowledge-augmented recommendation tasks. 
Specifically, we use the same categories' user-item interaction data from the Amazon Review dataset~\cite{ni-etal-2019-justifying} shown in Table~\ref{Table: recommendation_stats}. 
We split datasets into train/dev/test sets at a ratio of 8:1:1 and report averaged RMSE~(root mean square error) scores over five runs.

To fairly evaluate the KG for recommendations, we sample the sub-graph from the original KG where \textit{co-buy} pairs are simultaneously purchased by at least one user in the recommendation training set. 
The detailed statistics of the matched KG are in Table~\ref{tab:matched_kg_stats}.
The item coverage computes the percentage of the items in the recommendation dataset that are covered by the matched KG. 
Moreover, we also filter the matched KG with the threshold of 0.5 or 0.9 on \textit{plausibility} and \textit{typicality} scores to evaluate the effectiveness of the knowledge population. 
From Table~\ref{tab:matched_kg_stats}, we can observe the number of edges essentially reduces when the filters are applied, but the coverage of the items does not drastically drop. 

\begin{table}[t]
\tiny
\centering
\begin{tabular}{l|cc}
\toprule
  &  \multicolumn{1}{c}{Clothing}   & \multicolumn{1}{c}{Electronics} \\
\midrule
\# Users &  782,144 & 486,349  \\
\# Items &  18,042  &  6,166   \\
\# Interactions & 1,579,499 & 1,056,406 \\
Density &  0.011\%  & 0.035\% \\
\bottomrule
\end{tabular}
\caption{Statistics of the recommendation datasets.}
\label{Table: recommendation_stats}
\end{table}

\paragraph{Knowledge Representation.} As our constructed KG can be represented as the triplet $((p_1, p_2), r, e)$, where the head $(p_1, p_2)$ is the co-buy pair, the relation $r$ is from relations in Table~\ref{tab:prompts} and $e$ refer to generated tails.
To combine both structural and textual information from KG, we modify the original TransE model~\cite{bordes2013translating} to the following objective: 
{
\begin{align}
    \mathcal{L} =  \gamma + d( \frac{ \bm{p_1} + \bm{p_2}}{2} + \bm{r}, \bm{e}) -  d( \frac{\bm{p_1'} + \bm{p_2'}}{2} + \bm{r},\bm{e}) \nonumber
\label{eq:modifired_transE}
\end{align}
}
where $\gamma$ is a margin parameter, and $\bm{p_1}$, $\bm{p_2}$, $\bm{p_1'}$, $\bm{p_2'}$ are item embeddings for positive head $(p_1, p_2)$, and negative corrupted head $(p_1', p_2')$. Meanwhile, $\bm{r}$ is the relation embedding for the relation $r$, $\bm{e}$ is the embedding for the tail $e$, and the function $d$ is Euclidean distance. 
Moreover, the node embeddings for $e$ are initialized by Sentence-BERT~\cite{reimers-2019-sentence-bert} representations.
After training the modified TransE model, all the item embeddings $\mathbf{p}$ can be used as extra features to enhance recommendations.

\begin{table}[t]
\tiny
\centering
\setlength\tabcolsep{4.5pt}
\begin{tabular}{l|cc|cc}
\toprule
\multicolumn{1}{c|}{\multirow{2}{*}{Knowledge Graph}} & \multicolumn{2}{c|}{Clothing}      & \multicolumn{2}{c}{Electronics}                     \\
\multicolumn{1}{c|}{}                    & \# Edges  &  Coverage & \# Edges  &  Coverage \\
\midrule
 Matched Knowledge Graph                                    & 432,119      & 79.83\%          & 117,836       & 82.40\%  \\
$\thinspace$ + Plau. \textgreater 0.5                        & 323,263      & 79.83\%         & 78,908         & 82.40\%  \\
$\thinspace$ + Plau. \textgreater 0.5 and Typi. \textgreater 0.5         & 141,422     & 79.67\%  & 40,978        & 80.20\%  \\
$\thinspace$ + Plau. \textgreater 0.9                                   & 269,210     & 79.83\%   & 58,013       & 82.39\%  \\
$\thinspace$  + Plau. \textgreater 0.9 and Typi. \textgreater 0.9      & 103,262    & 79.36\%       & 27,288      & 76.94\%  \\
\bottomrule
\end{tabular}
  \vspace{-0.05in}
\caption{Details of matched KG subsets. ``Plau.'' means plausibility and ``Typi'' means typicality.}
\label{tab:matched_kg_stats}
\end{table}

\subsection{Experimental Results}

\paragraph{Baselines.} We adopt commonly-used NCF~\cite{he2017neural} and Wide\&Deep model~\citep{cheng2016wide} as our baselines. As our goal is to evaluate the effectiveness of features derived from KG, we leave advanced KG fusion methods, such as hyperedges or meta path-enhanced, to future work. 

\paragraph{Ablation Study.} 
We conduct two ablation studies to evaluate the effect of structural information provided by the co-buy pairs and the semantic information provided by the tails' text only. 
For the former, we train a standard TransE model solely on co-buy pairs to learn the graph embeddings of items. 
For the latter, for each item in the matched KG, we conduct average pooling of its neighbor tails' Sentence-BERT embeddings as its semantic representations. 
The experimental results are shown in Table~\ref{tab:rec_result}, and we have the following observations. 
First, the textual information contained in intentional assertions is useful for product recommendations. This can be testified as the W\&D model can perform better even when only features of the assertions are provided. 
Second, our KG, even before annotations and filtering, can produce better item embeddings than solely using the co-buy item graphs. As we can see, the performance of our matched KG is better than that of the co-buy pair graphs. 
Third, the two-step annotation and population indeed help improve the item embeddings for recommendations. The higher the scores are, the larger improvement the recommendation system obtains.

\begin{table}[t]
\tiny
\centering
\begin{tabular}{l|cc}
\toprule
\multicolumn{1}{c|}{Method}&  \multicolumn{1}{c}{Clothing}  & \multicolumn{1}{c}{Electronics} \\
\midrule

NCF~\citep{he2017neural} & 1.117  &   1.086 \\

W\&D~\citep{cheng2016wide} & 1.104  &  1.071\\

$\thinspace$  + Co-Buy Structure Only  & 1.096  &  1.067 \\
$\thinspace$  + Textual Features Only & 1.093  &  1.068 \\

$\thinspace$  + Matched Knowledge Graph & 1.093  & 1.058 \\
$\thinspace$  $\thinspace$ + Plau. > 0.5 &  1.087 & 1.060 \\
$\thinspace$ $\thinspace$ + Plau. > 0.5 and Typi. > 0.5 & \textbf{1.081} & 1.053 \\

$\thinspace$ $\thinspace$ + Plau. > 0.9 &  1.086 & 1.053 \\
$\thinspace$ $\thinspace$ + Plau. > 0.9 and Typi. > 0.9 &  \textbf{1.081} & \textbf{1.052} \\
\bottomrule
\end{tabular}
  \vspace{-0.05in}
\caption{Recommendation results in RMSE.}
\label{tab:rec_result}
\end{table}

\section{Related Work}
\label{appendix:relatedwork}

\noindent \textbf{Knowledge Graph Construction.}
An early approach of commonsense KG construction is proposed in ConceptNet~\cite{liu2004conceptnet,speer2017conceptnet} where both text mining and crowdsourcing are leveraged. 
In 2012, a web-scale KG, Probase, which focuses on {\it IsA} relations~\cite{yu2020hearst}, is constructed based on pattern mining~\cite{wu2012probase}, which can model both plausibility and typicality of conceptualizations~\cite{song2011concept}.
Recently, situational commonsense knowledge, such as Event2Mind~\cite{rashkin-etal-2018-event2mind} and ATOMIC~\cite{sap2019atomic}, has attracted more attention in the field of AI and NLP.
Then their extensions and neural generative models are developed~\cite{bosselut-etal-2019-comet,hwang2021comet}.
Meanwhile, information extraction can be used to extract event-related knowledge from large-scale corpora, such as KnowllyWood~\cite{TandonMDW15KnowlyWood}, WebChild~\cite{TandonMW17WebChild2}, and ASER~\cite{zhang2020aser,DBLP:journals/ai/ZhangLPKOFS22}.
The extracted knowledge can then be transferred to other human-annotated knowledge resources~\cite{DBLP:conf/ijcai/ZhangKSR20,DBLP:conf/www/FangZWSH21,DBLP:conf/emnlp/FangWCHZSH21}.

In e-commerce, Amazon Product Graph~\cite{zalmout2021all} is developed to align Amazon catalog data with external KGs such as Freebase and to automatically extract thousands of attributes in millions of product types~\cite{DBLP:conf/acl/KaramanolakisMD20,DBLP:conf/kdd/DongHKLLMXZZSDM20,DBLP:conf/www/ZhangZLDSF022}.
Alibaba also develops a series of KGs including AliCG~\cite{zhang2021alicg}, AliCoCo~\cite{luo2020alicoco,luo2021alicoco2}, AliMeKG~\cite{li2020alimekg}, and OpenBG~\cite{2022_OpenBG,qu2022commonsense}.
As we have stated in the introduction, there is still a gap between collecting factual knowledge about products and modeling users' purchasing intentions.

\noindent \textbf{Language Models as Knowledge Bases.} Researchers have shown LLMs trained on large corpus encode a significant amount of knowledge in their parameters~\cite{alkhamissi2022review,ye2022generative}. 
LLMs can memorize factual and commonsense knowledge, and one can use prompts~\cite{liu2021pre} to probe knowledge from them~\cite{petroni2019language}.
It has been shown that we can derive factual KGs at scale based on LLMs for factual knowledge~\cite{DBLP:journals/corr/abs-2010-11967,hao2022bertnet} and distill human-level commonsense knowledge from GPT3~\citep{west-etal-2022-symbolic}. 
None of the above KGs are related to products or purchasing intention.
We are the first to propose a complete KG construction pipeline from LLMs and several KG refinement methods for e-commerce commonsense discovery.

\section{Conclusion}

In this paper, we propose a new framework, \texttt{FolkScope}, to acquire intention commonsense knowledge for e-commerce behaviors. We develop a human-in-the-loop semi-automatic way to construct an intention KG, where the candidate assertions are automatically generated from large language models, with carefully designed prompts to align with ConceptNet commonsense relations.
Then we annotate both plausibility and typicality scores of sampled assertions and develop models to populate them to all generated candidates.
Then the high-quality assertions will be further structured using pattern mining and conceptualization to form more condensed and abstractive knowledge.
We conduct extensive evaluations to demonstrate the quality and usefulness of our constructed KG.
In the future, we plan extend our framework to multi-domain, multi-behavior type, multilingual~\cite{huang-etal-2022-multilingual,wang2023mutually} and temporal~\cite{wang2022rete,wanglearning} scenarios for empowering more e-commerce applications.

\section*{Limitations}

We outline two limitations of our work from \textit{user behavior sampling} and \textit{knowledge population} aspects. 
Due to huge-volume user behavior data produced every day in the e-commerce platform, it is crucial to efficiently sample significant behaviors that can indicate strong intentions and avoid random co-purchasing or clicking etc. 
Though in this work we adopt the criteria of selecting nodes whose degree are more than five in the \textit{co-buy} graph, it is still coarse-grained and more advanced methods remain to be explored in order to sample representative co-buy pairs for intention generation.
Some potential solutions are to aggregate frequent \textit{co-buy} category pairs and then sample product pairs within selected category pairs. 
Moreover, our proposed framework can be generalized to other types of abundant user behaviors such as \textit{search-click} and \textit{search-buy}, which requires to design corresponding prompts. 
We leave these designs to future work. 

For open text generation from LLMs, it becomes common practices to label high-quality data for finetuning to improve the quality and controllability of generation such as LaMDA~\cite{thoppilan2022lamda}, InstructGPT~\cite{ouyang2022training}, and ChatGPT\footnote{\url{https://openai.com/blog/chatgpt/}}. 
However, computation cost is the major bottleneck to use annotated data as human feedback for language model finetuning with billions of parameters, like \texttt{OPT-30b} in our work.
Hence we adopt a trade-off strategy to populate human judgements by training effective classifiers and conducting inferences over all the generation candidates. 
With impressive generation performance of ChatGPT, we expect efficient methods to directly optimize LLMs with human feedback in more scalable way like reinforcement learning~(RLHF), and enable LLMs to generate more typical intention knowledge with less annotation efforts.

\section*{Ethics Statement}

As our proposed framework relied on large language models, text generation based on LLMs often contains biased or harmful contexts.
We argue that our work largely mitigated the potential risks in the following ways. First, our careful-designed prompting leads to rather narrow generations constrained on small domains, i.e., products in e-commerce.
Second, we also had a strict data audit process for annotated data from annotators and populated data from trained classifiers.
On a small scale of inspections, we found none belongs to significant harmful contexts. 
The only related concern raised here is that some generated knowledge is irrelevant to the products themselves. The major reason is due to imprecise product titles written by sellers for search engine optimization, such as adding popular keywords to attract clicks or purchases.
Our human-in-the-loop annotation identified such cases and the trained classifier further assisted machines in detecting bias, as we hope our intention generations can be safe and unbiased as much as possible.   

\section*{Acknowledgements}

The authors of this paper were supported by the NSFC Fund~(U20B2053) from the NSFC of China, the RIF~(R6020-19 and R6021-20) and the GRF~(16211520 and 16205322) from RGC of Hong Kong, the MHKJFS~(MHP/001/19) from ITC of Hong Kong and the National Key R\&D Program of China~(2019YFE0198200) with special thanks to HKMAAC and CUSBLT. We also thank the support from the UGC Research Matching Grants~(RMGS20EG01-D, RMGS20CR11, RMGS20CR12, RMGS20EG19, RMGS20EG21, RMGS23CR05, RMGS23EG08). 

\bibliography{acl2023}
\bibliographystyle{acl_natbib}

\newpage

\appendix
\section*{Appendix}

\section{Annotation Guideline}
\label{sec:annotation_detail}

Workers satisfying the following three requirements are invited to participate: (1) at least 90\% lifelong HITs approval rate, (2) at least 1,000 HITs approved, and (3) achieving 80\% accuracy on at least 10 qualification questions, which are carefully selected by authors of this paper. Qualified workers will be further invited to annotate 16 tricky assertions. Based on workers' annotations, they will receive personalized feedback containing explanations of the errors they made along with advice to improve their annotation accuracy. Workers surpassing these two rounds are deemed qualified for main-round annotations. To avoid spamming, experts will provide feedback for all workers based on a sample of their main rounds' annotations from time to time. Finally, we recruited more than 100 workers in the \texttt{us-east} district. It takes \$0.2 on average for each assertion, and the annotators are paid \$7.7 per hour on average, which satisfies the local minimum wage under local laws.

\begin{figure}[t]
      \centering
      \includegraphics[scale=0.35]{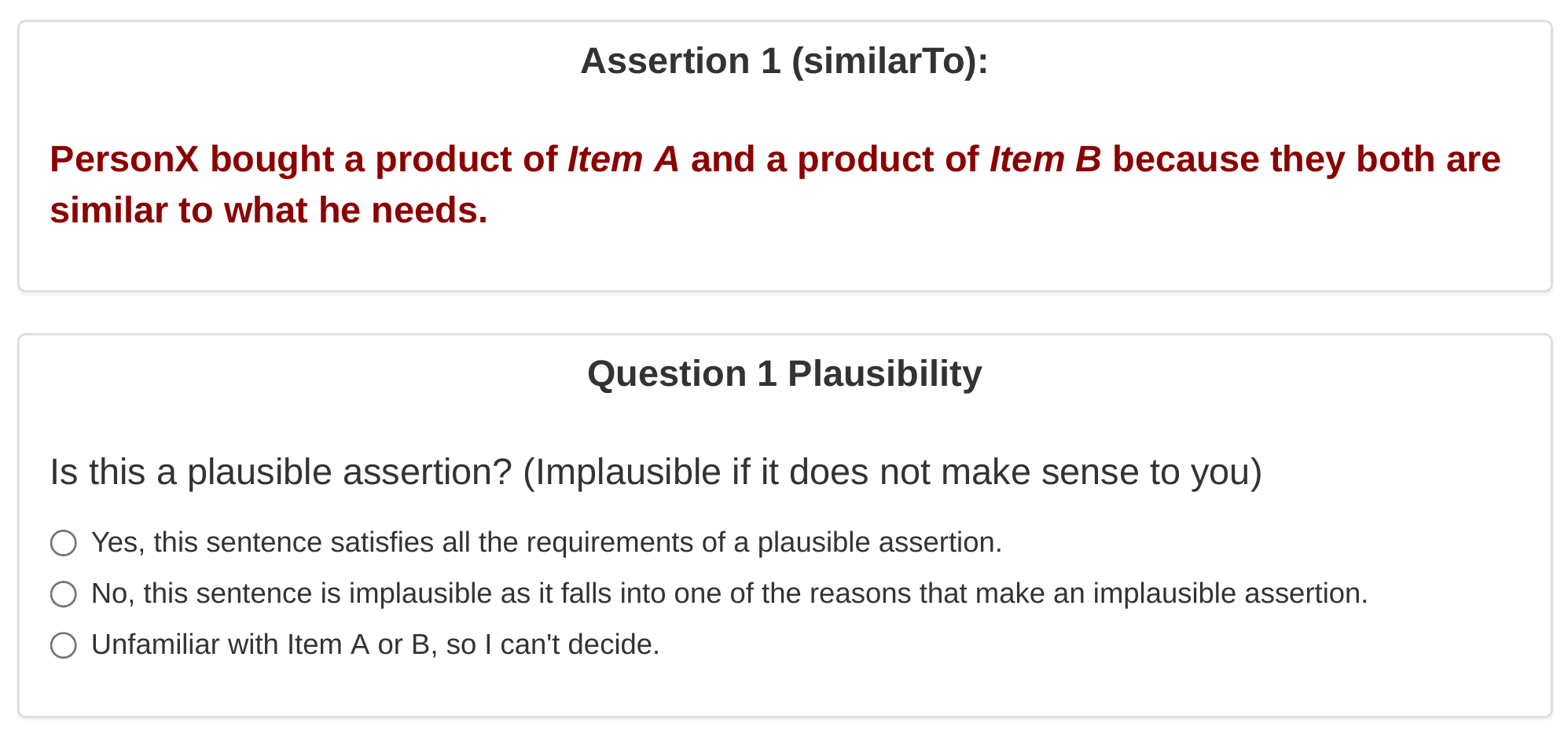}
      \caption{The question card in our plausibility annotation round.}
      \label{fig:validity_question}
\end{figure}

\begin{figure}[t]
      \centering
      \includegraphics[scale=0.35]{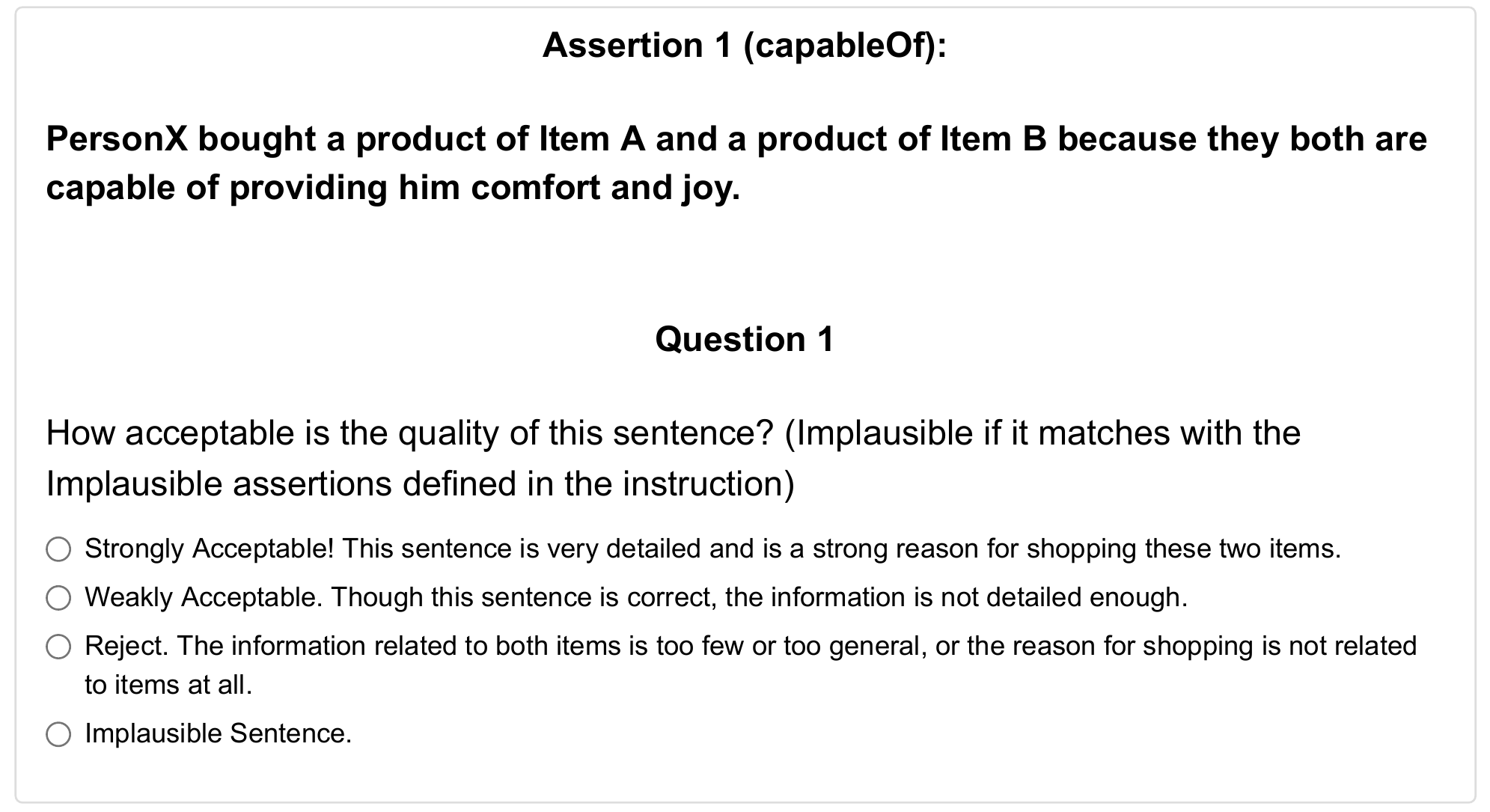}
      \caption{The question card in our typicality annotation round.}
      \label{fig:quality_question}
\end{figure}

We conducted human annotations and evaluations on the Amazon Mechanical Turk as Figure~\ref{fig:validity_question} for the first-step plausibility
annotation and as Figure~\ref{fig:quality_question} for the second-step typicality annotation.

\section{Knowledge Population}\label{appendix:KBP}

\begin{figure}
     \centering
     \includegraphics[scale=0.4]{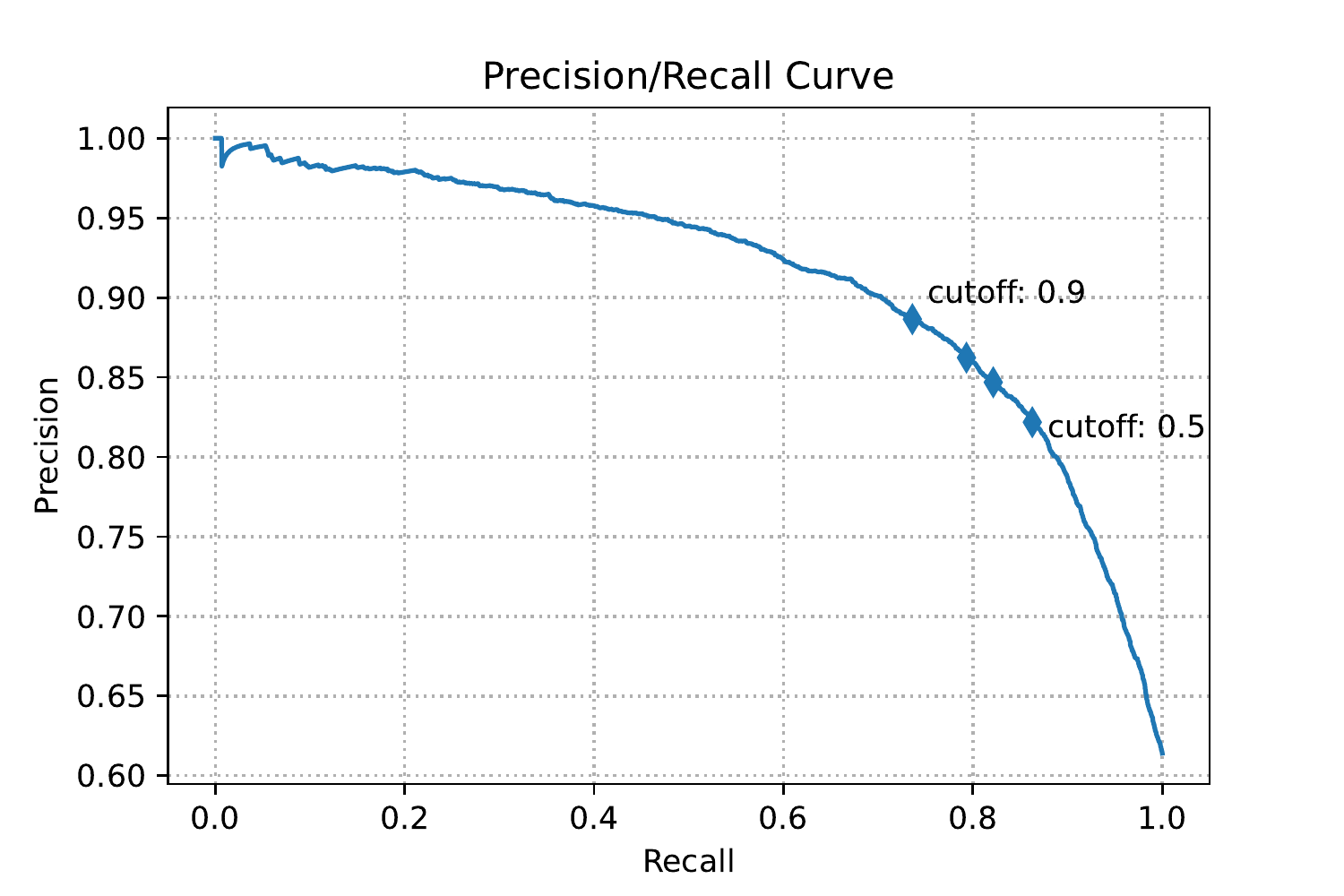}
     \caption{The precision-recall curve of our plausibility population classifier on the human-labeled validation set. The annotated points show the different thresholds~(cutoffs) to filter the generated assertions, i.e. from left to right: 0.9, 0.8, 0.7, 0.5 respectively.}
     \label{fig:prcurve}
\end{figure}

Using different confidence cutting-off thresholds leads to trade-offs between the accuracy of generation and the size of the corpus. 
Higher values result in conservative selections that favor precision over recall, whereas lower ones tend to recall more plausible assertions.
We plotted four cutoff points in Figure~\ref{fig:prcurve}.

\section{Pattern Mining Details} \label{appendix:pattern_mining}


We apply the frequent graph substructure mining algorithm over dependency parse trees to discover the linguistic patterns.
We sample 90,000 candidates for each relation to analyze patterns and then parse each candidate into a dependency tree.
In addition, the lemmatized tokens, pos-tags, and named entities are acquired for further use.
To reduce the time complexity of pattern mining, we mine high-frequency patterns for each relation.
To meet the two requirements of the knowledge with high precision but non-trivial, patterns are required to perfectly match more than 500 times.
One perfect match means that this pattern is the longest pattern, and no other candidate patterns can match.
Therefore, the pattern mining pipeline consists of three passes: (1) a graph pattern mining algorithm, Java implementation of gSpan~\cite{DBLP:conf/icdm/YanH02},\footnote{\url{https://github.com/timtadh/parsemis}} to mine all candidate patterns with the frequency more than 500, (2) a subgraph isomorphism algorithm, C++ implementation of VF2 algorithm in igraph,\footnote{\url{https://igraph.org/}} with a longest-first greedy strategy to check the perfect match frequency, and (3) human evaluation and revision.
Finally, we obtain 256 patterns that cover 80.77\% generated candidates.
Details can be found in Table~\ref{tab:coverage}.

\begin{table}\small
\centering
\setlength\tabcolsep{4pt}
\begin{tabular}{l|c|c|c}
     \toprule
     Type & Relation & \# of Patterns & Coverage \\
     \midrule
     \multirow{10}{*}{\begin{minipage}{0.4in}Item\end{minipage}}& \textit{RelatedTo} & 14 & 96.94 \\
     & \textit{IsA} & 15 & 97.20 \\
     & \textit{HasA} & 12 & 99.30 \\
     & \textit{PartOf} & 8 & 99.83 \\
     & \textit{MadeOf} & 13 & 99.45 \\
     & \textit{SimilarTo} & 7 & 22.22 \\
     & \textit{CreatedBy} & 14 & 98.59 \\
     & \textit{HasProperty} & 16 & 63.20 \\
     & \textit{DistinctFrom} & 9 & 97.30 \\
     & \textit{DerivedFrom} & 20 & 100.00 \\
     \midrule
     \multirow{5}{*}{\begin{minipage}{0.4in}Function\end{minipage}}& \textit{UsedFor} & 2 & 96.57 \\
     & \textit{CapableOf} & 13 & 74.68 \\
     & \textit{DefinedAs} & 27 & 95.99 \\
     & \textit{SymbolOf} & 9 & 99.76\\
     & \textit{MannerOf} & 34 & 98.56 \\
     \midrule
     \multirow{3}{*}{\begin{minipage}{0.4in}Human\end{minipage}}& \textit{Cause} & 21 & 93.68 \\
     & \textit{Result} & 0 & 0 \\
     & \textit{CauseDesire} & 0 & 0 \\
     \midrule 
     \textit{Overall} & / & 256 & 80.77 \\
     \bottomrule
\end{tabular}
\caption{Frequent linguistic patterns and corresponding coverage on human-annotated knowledge.}\label{tab:coverage}
\end{table}

\section{More Evaluations}\label{appendix:more_evaluation}
\subsection{Implicit Generation Evaluation}

As we know, language model based generation capture spurious correlation given the condition of the generation~\citep{ji2022survey}.
Hence we simply quantify the diversity as the novelty ratio of generated tails not appearing in the item titles, i.e., novel generations.
Different from explicit attribute extraction~\cite{vilnis2022impakt,yang2022mave}, our generative method is able to extract implicit knowledge behind item titles or descriptions.  
For example, the title ``Diesel Analog Three-Hand - Black and Gold Women's watch'' contains specific attributes like ``Black and Gold'' or type information ``women's watch.'' Such knowledge can be easily extracted by off-the-shelf tools. 
Traditional information extraction based approaches mostly cover our knowledge if the generation simply copies titles to reflect the attributes.
Otherwise, it means that we provide much novel and diverse information compared with traditional approaches.
The novelty ratio increases from 96.85\% to 97.38\% after we use the trained classifiers for filtering. Intuitively, filtering can improve the novelty ratio.
For the assertions whose typicality scores are above 0.9, we also observe that the novelty ratio reaches 98.01\%. 
These findings suggest that \texttt{FolkScope} is indeed an effective framework for mining high-quality implicit knowledge.

\subsection{Fine-grained Subcategory Knowledge}
Since the items are organized in multilevel fine-grained subcategories in the catalog of shopping websites, we are interested in whether our constructed KG contains high-quality common intentions among items belonging to subcategories.
The common knowledge can be useful to have intention-level organizations besides category-level and further help downstream tasks.  
The co-buy item-pairs in our sampled \textit{clothing} category fall into 15,708 subcategory pairs, such as (\textit{necklaces}, \textit{earning}) or (\textit{sweater}, \textit{home \& kitchen}), where most of them are different subcategories in one pair.
We select frequent common assertions with high typicality scores to demonstrate the abstract knowledge. Two examples are shown in Table~\ref{tab:subcategory_case}.
Though costumes and toys belong to two different types, they are complementary because of the same usage, such as  ``Halloween,'' ``Easter holiday,'' and ``Christmas,'' or sharing the same key feature like ``star war character,'' ``pirate.''
On the other hand, if two items fall in the same subcategory, like ``dresses'' in Table~\ref{tab:subcategory_case}, 
the generated assertions share some common characteristics, such as being suitable for certain events and complementing each other when worn together.

\begin{table}[t]\tiny
\centering
\setlength\tabcolsep{3pt}
\begin{tabular}{l|c}
\toprule
Subcategory  & Generation         \\ \midrule 
\multirow{6}{*}{\begin{minipage}
{0.6in}{(Costumes, Toys)}\end{minipage}}
& he wants to disguise himself as a superhero  \\
& they can be used to make a crown costume    \\
& he wanted to be a star war character for Halloween \\
& they are both a manner of Christmas decoration  \\
& he wants kids to have fun and enjoy the Easter holiday \\
& he is able to dress up as a pirate \\
\midrule
\multirow{6}{*}{\begin{minipage}{0.6in}(Dresses, Dresses)\end{minipage}}
& they are symbol of the fashion trend  \\
& they can both be worn to formal events    \\
& they can both be worn for casual occasions \\
& they are both used for wedding dress  \\
& they are both capable of giving a good fit \\
& they can both being worn by girls of any age \\
\bottomrule
\end{tabular}
\caption{Generated knowledge in same subcategory.
}\label{tab:subcategory_case}
\end{table}

\subsection{Use Different LLMs as Knowledge Source}

We are interested in whether different sizes of language models have large impact on the generation. 
Hence we empirically analyze the plausible rate of generation using four language models: \texttt{GPT-J}~(6b), \texttt{OPT-30b}, \texttt{OPT-66b} and \texttt{text-davinci-003}.
We can observe that: 1) \texttt{OPT-30b} outperformed \texttt{GPT-J} over 10\% (51\% vs. 41\%) while \texttt{OPT-66b} did not improve \texttt{OPT-30b}.
2) \texttt{text-davinci-003} achieved nearly perfect results and make little mistakes when recognizing products given title information.
Though impressive results, we have to balance between knowledge size and cost hence the takeaway from our work is to use human annotation with middle-size LLMs.

\begin{table*}[b]\small
\centering
\setlength\tabcolsep{3pt}
\begin{tabular}{l|c|c}
\toprule
Relation  & Clothing  & Electronics         \\ \midrule
\multirow{3}{*}{\begin{minipage}{0.7in}Open\end{minipage}}
& a fan of Harry Potter / Star Wars & make a robot, make a remote control, build a PC  \\
& give gifts for his girlfriends / his son &  know how to play guitar / take better photos    \\
& go to a costume party / wedding / be a father & learn code / microcontroller programming  \\
\midrule 
\multirow{6}{*}{\begin{minipage}
{0.7in}{UsedFor}\end{minipage}}
& outdoor activities, hiking, camping, travel & outdoor use, navigation, education, networking \\
& daily use, formal occasions, winter sports & personal use, office work, home theater, 3D movies  \\
& babies, maternity wear, sleepwear & baby photography, underwater photography   \\
& Halloween costumes, Christmas cosplay & Arduino projects, Raspberry Pi, Samsung headphone \\
& jewelry making, leather care, weight loss & water cooling, cable management, screen protection\\
& nursing, working out, polishing shoes & framing, storing data, mounting camera, prototyping \\ 
\midrule
\multirow{5}{*}{\begin{minipage}{0.7in}CapableOf\end{minipage}}
& keeping cool, keeping dry, keeping warm & taking pictures, printing labels, boosting signals \\
& being worn with jeans / dress / shorts & being used in car / boat / computer / water / emergency \\
& holding up pants, holding a lot of stuff & holding radio / CDs / GoPro camera / phones / devices \\
& protecting from rain / sun / harmful UV rays & tracking location / heart beat rate / cycling activities \\
& making him look like wizard / price / Batman & controlling light / TV / home automation / device  \\
\midrule
\multirow{3}{*}{\begin{minipage}{0.7in}SymbolOf\end{minipage}}
& his love for daughter / wife / mother / family & his passion for gaming / aviation /cycling / sports \\
& luxury, friendship, childhood, the 80s  & security, reliability, durability, high performance   \\
& modern life,  American culture, graduation  & latest technology, hacker culture, music industry \\
\bottomrule
\end{tabular}
\caption{More examples of high-frequency typical assertions for different relations.
Note we omit the prompts for space and simplicity .}\label{tab:more_typical_examples}
\end{table*}


\end{document}